\documentclass[preprint,review,12pt]{elsarticle}

\usepackage{amssymb}
\usepackage{url}            
\usepackage{booktabs}       
\usepackage{amsfonts}       
\usepackage{nicefrac}       
\usepackage{microtype}      
\usepackage{xcolor}         
\usepackage{multirow}
\usepackage{makecell}
\usepackage{graphicx}
\usepackage{bm}
\usepackage{amsmath}
\usepackage{subfigure}
\usepackage{amsthm,amsmath,amssymb}
\usepackage{mathrsfs}
\usepackage{algorithm}
\usepackage{algorithmic}
\usepackage{wrapfig}
\usepackage{natbib}
\makeatletter
\newcommand*\bigcdot{\mathpalette\bigcdot@{.5}}
\newcommand*\bigcdot@[2]{\mathbin{\vcenter{\hbox{\scalebox{#2}{$\m@th#1\bullet$}}}}}
\makeatother

\journal{Neural Networks}

\begin{document}

\begin{frontmatter}

\title{Object-fabrication Targeted Attack for Object Detection}

\author{Xuchong Zhang\corref{cor1}}
\author{Changfeng Sun\corref{ss}}
\author{Haoliang Han}
\author{Hongbin Sun}

\cortext[cor1]{Corresponding email address: zhangxc0329@xjtu.edu.cn}
\cortext[ss]{First author and second author contribute equally to this work.}

\address{ Institute of Artificial Intelligence and Robotics, Xi’an Jiaotong University, Shaanxi, China}

\begin{abstract}
Recent studies have demonstrated that object detection networks are usually vulnerable to adversarial examples. Generally, adversarial attacks for object detection can be categorized into targeted and untargeted attacks. Compared with untargeted attacks, targeted attacks present greater challenges and all existing targeted attack methods launch the attack by misleading detectors to mislabel the detected object as a specific wrong label. However, since these methods must depend on the presence of the detected objects within the victim image, they suffer from limitations in attack scenarios and attack success rates. In this paper, we propose a targeted feature space attack method that can mislead detectors to `fabricate' extra designated objects regardless of whether the victim image contains objects or not. Specifically, we introduce a guided image to extract coarse-grained features of the target objects and design an innovative dual attention mechanism to filter out the critical features of the target objects efficiently. The attack performance of the proposed method is evaluated on MS COCO and BDD100K datasets with FasterRCNN and YOLOv5. Evaluation results indicate that the proposed targeted feature space attack method shows significant improvements in terms of image-specific, universality, and generalization attack performance, compared with the previous targeted attack for object detection.
\end{abstract}


\end{frontmatter}
\section{Introduction}
\label{intro}
Object detection~\cite{survey_objectdetection} has achieved remarkable performance by employing deep neural networks and been applied to many applications such as autonomous driving~\cite{od-autodriving,ob-autodriving2}, intelligent surveillance~\cite{mishra2016study,kim2010intelligent}, and mobile robot~\cite{DBLP:journals/ral/KimKLL22,DBLP:conf/iccais/LiXWZLC21}.
However, a variety of researches~\cite{szegedy-et-al:intriguing,DBLP:journals/nn/HuSYJTCL24,DBLP:journals/nn/HuSJYTCZL24} have shown that the detection network is vulnerable to adversarial attack and usually gives wrong predictions when encountering adversarial examples crafted by adding human-imperceptible perturbations to input images.
Due to the great threat on the applications of object detection, adversarial attacks have gained widespread attention.

\begin{figure}[t]
\setlength{\abovecaptionskip}{-0.2cm}
\centering
\includegraphics[width=0.8\linewidth]{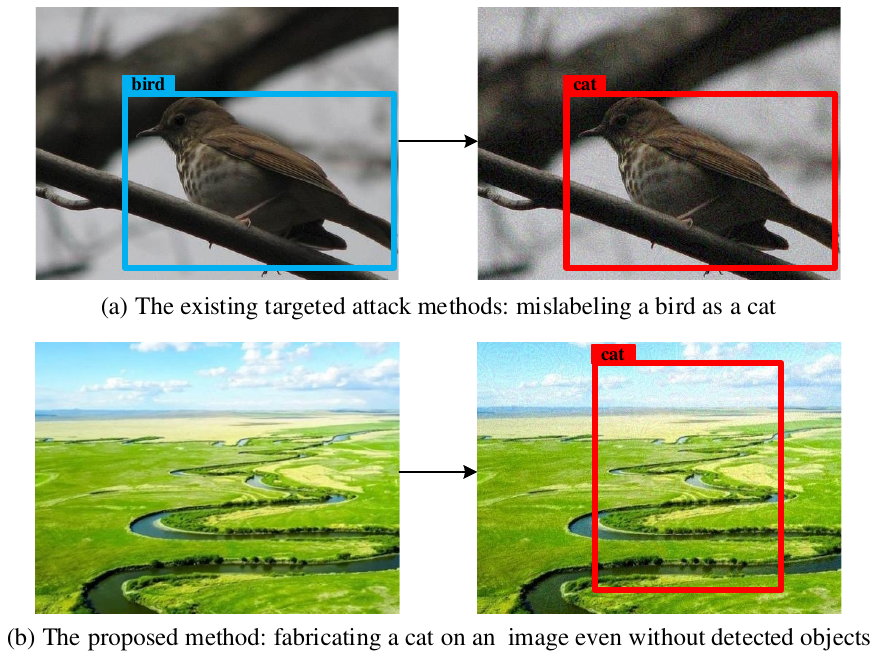}
\caption{  The existing targeted attack methods VS the proposed targeted feature space attack. The existing methods must rely on the detected objects to launch attacks while the proposed method can launch attacks on any image regardless of whether contains objects or not.}
\label{fig.intro}
\end{figure}

Generally, adversarial attacks for object detection can be grouped into untargeted attacks and targeted attacks. The former aims to mislead the detectors to predict objects to other arbitrary labels or none labels~\cite{dfool, ca}, while the latter is used to fool the detectors to predict certain specific wrong labels~\cite{cai, chow-et-al:tog}. Compared with the untargeted attack, the targeted attack for object detection is a more challenging task and there are relatively few methods designed for it. As shown in Fig.~\ref{fig.intro}(a), current targeted attack methods typically attempt to deceive detectors by mislabeling an existing object as a specific wrong label. They add human-imperceptible perturbations to the victim images to change the features of the victim objects, consequently misleading the detectors into predicting the victim object as the designated incorrect label. Since the existing methods must rely on the detected objects to launch the attack, they have several limitations. 
Firstly, the attack cannot be carried out on images that do not contain any objects, and the attack tends to fail when the designated label's features differ significantly from those of the victim object. 
Additionally, due to the victim objects having different scales, colors, positions, and features across different images, it is difficult for the existing attack method to generate a universal perturbation that can simultaneously mislabel the victim objects in different images. 
Consequently, the perturbation must be retrained for each new image.
Given these limitations, there is significant value in developing a more flexible targeted attack method that does not rely on the presence of detected objects.

As illustrated in Fig.~\ref{fig.intro}(b), a more flexible and effective target attack method should be capable of misleading detectors into predicting the target object in any images, regardless of whether they contain objects or not. 
A natural approach to achieve this is to construct the target object's semantic features and implant them into the victim images through perturbations.
However, since neural networks interpret features very differently from humans, it is not feasible to manually design the target object's semantic features. 
Although it is well known that features can be represented by the output of certain layers in detection networks, using features extracted from images containing the target objects is also not ideal. 
This is because such features are often mixed with irrelevant information, such as background elements or other objects, which leads to confusion.
Therefore, precisely constructing the target object's semantic features is a challenging task. Besides this, to ensure the stealthiness of the attack, the perturbation should be imperceptible to humans. 
Designing small, imperceptible perturbations that accurately approximate the target features and implant them into victim images further complicates this task.

In this paper, to address the above issues, we propose a targeted feature space attack (TFA) method for object detection to mislead the detectors to predict target objects on any images regardless of whether they contain objects or not.
Firstly, we introduce a guided image containing the target objects to extract the coarse-grained features of the target objects.
Then, we design a dual attention mechanism to efficiently filter out the critical features of the target object from the coarse-grained features. 
Specifically, spatial attention is used to determine the spatial location of the target object, and class attention is used to select the most critical channels that are associated with the semantic information of the target object.
Finally, a loss function is designed to approximate the target feature by minimizing the distance between the filtered guided image feature and the adversarial example feature, and the perturbation is obtained by optimizing the designed loss.
In order to evaluate the proposed TFA method, we conduct extensive experiments on MS COCO and BDD100K datasets to attack two representative object detection models, i.e. FasterRCNN~\cite{ren-et-al:fasterrcnn} and YOLOv5~\cite{yolov5}. Because the existing methods are inappropriate for the object-independent attack, we implement a variant of the previous method~\cite{chow-et-al:tog,cai}  to compare the performance with TFA.
The experimental results clearly demonstrate that the proposed TFA method achieves significant improvements in the cases of image-specific attack success rates, universal and generalization performance.

Accordingly, the main contributions of our work are described as follows.

\begin{itemize}
	\item  Compared with the existing targeted attack methods which mislabel detected objects as specific wrong labels, we propose a more flexible and effective attack method that can mislead detectors to ‘fabricate’ extra designated objects with specific target labels. 
 
    \item We further design a dual attention mechanism to construct the target object features and implement the targeted feature space attack. To the best of our knowledge, our work is the first to employ the feature space attack for the targeted attack of object detection.

    \item The experiments on multiple datasets and detectors clearly show that the proposed TFA method achieves significant improvements over the existing SOTA targeted attack method in terms of image-specific, universality, and generalization attack performance.
 \end{itemize}

\section{Related Work}
\label{sec:relate}
The adversarial example~\cite{goodfellow2014explaining,fia,DBLP:journals/nn/LiHY24} was originally proposed to disrupt an image classifier by adding a small perturbation that is imperceptible by humans to the input images. Then, the researches of the adversarial example were extended to other tasks~\cite{adv_makeup,leida,Tu2020,DBLP:journals/nn/WangLSL24} and gradually evolved into two types of methods, i.e. untargeted attack and targeted attack. This paper mainly focuses on the targeted attack for object detection. 
Compared with the image classification task, attacking object detection is more complex because the attacks on detectors need to take two types of predictions into account, i.e. presence/absence of an object and the class of the object~\cite{du2022detects}.

The goal of the targeted attack is to fool the detectors to detect a specific wrong label. Based on our extensive investigations, the research on targeted attacks is relatively fewer than untargeted attacks. Xie \emph{et.al.}~\cite{xie-et-al:dag} propose the first targeted method named dense adversary generation (DAG), which assigns a specific adversarial label to the proposals, then designs a loss function to maximize the classification score of that label and finally performs iterative gradient backpropagation to purposely mislabel the proposals.
Based on DAG, Nezami \emph{et.al.}~\cite{Nezami-et-al:pick} attack a designated object in an image and preserve the labels of other detected objects in the image by adding large perturbations to the bounding box of the designated object.
Chow \emph{et.al.}~\cite{chow-et-al:tog} propose a targeted adversarial objectness gradient attack (TOG). By adding human-imperceptible perturbations to victim images, TOG attempts to improve the classification score of a specific class and reduce the classification score of an initially detected object, thus achieving object-mislabeling attack.
Based on TOG, Cai \emph{et.al.}~\cite{cai} proposed CAA method by introducing joint training strategy and context information to mislabel detected objects as the specific wrong labels and enhance the transferability of the adversarial examples. In these methods,~\cite{xie-et-al:dag} and~\cite{Nezami-et-al:pick} are designed for attacking only two-stage detectors, while TOG and CAA are appropriate for both one-stage and two-stage detectors. Besides, TOG and CAA also achieve remarkable targeted attack performance.

The detection network is usually updated by optimizing the loss function consisting of bounding boxes regression, objectness measures, and class probability vectors. 
We can see that all the existing targeted attack methods launch the attack by mislabeling the detected objects. 
They all attack the final output layer to achieve the mislabeling by maximizing the detected object's classification score of the specific wrong label.
Therefore, these methods cannot be directly applied to the images without any objects since they need detectors to first predict an object, and then can then launch attacks by maximizing the classification score of that object. 
As shown in the introduction section, the reliance on the detected objects brings several shortcomings and motivates us to design an efficient method that can launch attacks on any image regardless of contains objects or not.

\section{Method}
\label{sec:TFA}
\subsection{Overall Framework}
In order to ‘fabricate’ a fake object in the victim image without relying on the detected object, we propose to attack the internal feature layer of the detection model instead of the final output layer which is exploited in traditional targeted attacks. The key idea of the proposed method is to employ the targeted feature space attack to change part of the semantic feature of the victim image towards that of the target object in the guided image, thus fooling the detector into predicting the target object with a specific label in the victim image.

\begin{figure*}[t]
\setlength{\abovecaptionskip}{-0.2cm}
	\centering
	\includegraphics[width=0.9\columnwidth]{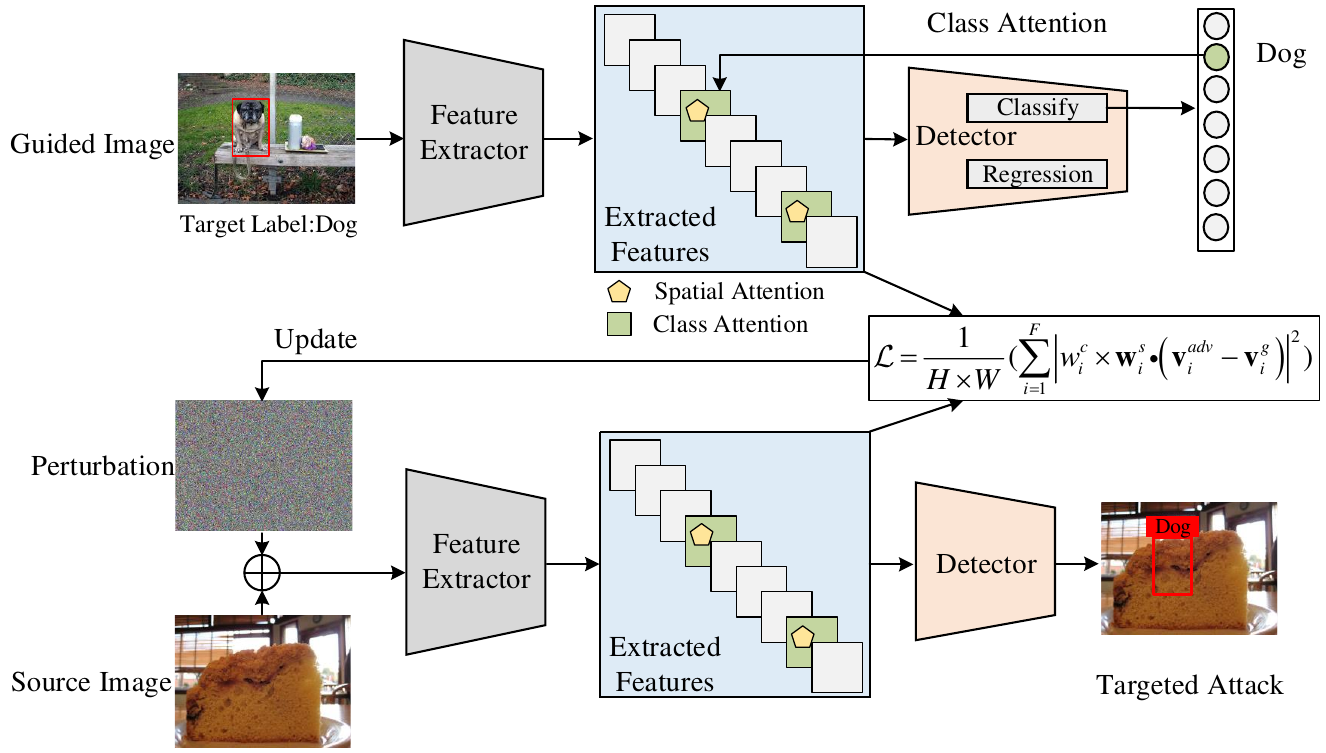}
	\caption{The framework of the proposed targeted feature space attack.  A victim image $\textbf{x}^s$ and a guided image $\textbf{x}^g$ containing a target object $o^t$ are used to conduct the targeted attack. Firstly, the features of the guided image and adversarial example are extracted by two identical detection models, i.e. $\textbf{v}^{g}_i$ and $\textbf{v}^{adv}_i$.
	Then, the critical features that represent $o^t$ are filtered out by the proposed dual attention mechanism. Finally, the adversarial example can be obtained by optimizing a loss function $\mathcal{L}$ consisting of the attention weights (${w}^c_i$, $\textbf{w}^s_i$) and features ($\textbf{v}^{g}_i$, $\textbf{v}^{adv}_i$). }
	\label{fig.overall}
\end{figure*}

The overall framework of the proposed TFA is shown in Fig.~\ref{fig.overall}. Firstly, a guided image is introduced to construct the target object feature.
Next, a well-trained object detection model is applied to extract the features of the guided image and the victim image.
Then, a dual attention mechanism is employed to filter out the critical features that can represent the target object (e.g. dog) in the extracted features of the guided image. 
After that, the small perturbation is constantly updated by optimizing the loss function consisting of the guided image feature space and adversarial example feature space.
Finally, the final adversarial example for the specific class can be obtained by adding the perturbation to the victim image. 
In the following content, we will formulate the proposed targeted attack method and present the attack process in detail.

\subsection{TFA Formulation}
Given an input victim image $\textbf{x}^s$, an object detector $G$ without suffering from any attacks is able to correctly detect the objects $\textbf{O}^s$ that are originally contained in the image, i.e. $ G\left(\textbf{x}^s\right) = \textbf{O}^s$, where $\textbf{O}^s$ is a small subset of the whole object dataset $\{o_1, o_2, o_3, ..., o_n\}$. The goal of the proposed targeted attack method is to find adversarial examples $ \textbf{x}^{adv} = \textbf{x}^s + \bm{\delta} $ such that $G\left(\textbf{x}^{adv}\right)$ can yield a new target object $o^t$ ($o^t \notin \textbf{O}^s$) on a certain region of $\textbf{x}^s$.  
$\bm{\delta}$ is an imperceptible perturbation which should be limited to a small range.
In this paper, we use the $\ell_\infty$ norm to restrict the value of $\bm{\delta}$ less than $ \epsilon$, i.e., $ \Vert \bm{\delta} \Vert_\infty \leq \epsilon$. Therefore, the optimization of $\textbf{x}^{adv}$ for the targeted attack is formulated as,
\begin{equation}
\begin{aligned}
\textbf{x}^{adv}=\mathop{\arg\min}\limits_{\textbf{x}} \mathcal{L} \left( 
G \left( \textbf{x}\right) , o^t\right), s.t. \Vert \textbf{x}^{adv}- \textbf{x}^s \Vert_\infty \leq \epsilon,
\end{aligned}
\label{e1}
\end{equation}
where $\mathcal{L} \left(G \left( \textbf{x}\right), o^t\right)$ is the loss function constructed by the feature spaces of the guided image and adversarial example. It is calculated as,
\begin{equation}
\begin{aligned}
\mathcal{L} = \frac{1}{H \times W}\left(  \sum_{i=1}^F\left|  \textbf{w}_i \bigcdot  \left( \textbf{v}^{adv}_i - \textbf{v}_i^g \right)\right|^2\right), 
\end{aligned}
\label{e2}
\end{equation}
where $\textbf{v}^{adv}_i$ denotes the $i$-th  channel of the extracted features of the adversarial example $ \textbf{x}^{adv}$, $\textbf{v}^g_i$ denotes the $i$-th  channel of the extracted features of the guided image $\textbf{x}^g$, $F$ is the total number of the feature channels, $H$ and $W$ are the horizontal and vertical resolutions of the feature map, $\textbf{w}_i$ is the attention weight to differentiate the importance of different features and more specific descriptions are presented in the next subsection.
Based on Equation (\ref{e1}) and (\ref{e2}), we can conclude that the adversarial example is optimized by gradually driving the corresponding features of the victim image towards the critical features of the target object in the guided image.

\subsection{Class Attention and Spatial Attention}
\label{subsec:dual_att}
As mentioned above,  filtering out the key features of the target object in the whole feature space is critical to achieve the proposed targeted attack. Hence, we propose a dual attention mechanism to address this problem, which includes class attention and spatial attention.

The class attention is used to select the most critical channels in the extracted features which are closely associated with the semantic information of the target object. 
Inspired by~\cite{Selvaraju-et-al:gradcam}, we use the derivative of the objectness with respect to the extracted features $\textbf{v}^g$ to indicate the contribution of different channels to	the category classification of the target object, thus these different contributions can be calculated and regarded as the class attention weight $ \textbf{w}^c$. Specifically, it is formulated as the global-average-pooling over the width and height dimensions.
\begin{equation}
\begin{aligned}
\hat{w}_i^c =  \cfrac{1}{H \times W} \sum_m \sum_n  \cfrac{\partial y^t}{\partial \textbf{v}^g_{imn}},
\end{aligned}
\label{e3}
\end{equation}
where $y^t$ is the confidence score of the specific class $t$,
$\hat{w}_i^c$ is the class weight of the $i$-th channel, and $i,m,n$ in $\textbf{v}^g_{imn}$ are the indices of the feature channel,  width and height dimensions respectively.  It should be noted that $\hat{w}_i^c$  has positive and negative values, where the former means the $i$-th feature channel has a positive correlation with the class of the target object, and the latter means the $i$-th feature channel has a negative correlation with the class of the target object. We only keep the positive values, thus the final class attention weight $ {w_i}^c$ is calculated as, 
\begin{equation}
\begin{aligned}
w_i^c = max\left(\hat{w}_i^c , 0 \right) =max\left( \cfrac{1}{H \times W} \sum_m \sum_n  \cfrac{\partial y^t}{\partial \textbf{v}^g_{imn}},0\right).
\end{aligned}
\label{e4}
\end{equation}
A large value of $w_i^c$  means that the $i$-th channel is more relevant to the semantic feature of the target object.

The spatial attention is used to determine the spatial location of the target object in feature space.  
As shown in~\cite{zhou-et-al:ode}, the activation values at the locations containing objects are usually much larger than those containing the background in the feature map. Therefore, we define the positive values of the feature map $\textbf{v}^g_i$ as the spatial attention weight, 
\begin{equation}
\begin{aligned}
\textbf{w}^s_i  = max\left( \textbf{v}^g_i, 0 \right).
\end{aligned}
\label{e5}
\end{equation}
A large value in $\textbf{w}^s_i$ means that the corresponding position has a higher probability of being an object than the background.

After obtaining the class attention weight and the spatial attention weight, the final dual attention weight is defined as,
\begin{equation}
\begin{aligned}
\textbf{w}_i  ={w}^c_i \times \textbf{w}^s_i.
\end{aligned}
\label{e6}
\end{equation}

\subsection{Attack Algorithm}
According to Equation (\ref{e2})--(\ref{e6}), the final loss function of the proposed targeted attack is presented as,
\begin{equation}
\begin{aligned}
\mathcal{L} = \frac{1}{H \times W}\left(  \sum_{i=1}^F\left|  {w}^c_i \times \textbf{w}^s_i \bigcdot  \left( \textbf{v}_i^{adv} - \textbf{v}_i^g \right)\right|^2\right).
\end{aligned}
\label{e7}
\end{equation}
It is necessary to reiterate that $\textbf{v}^{adv}$ is the extracted features of the adversarial example, i.e. $(\textbf{x}^s + \bm{\delta})$.
Then, we use an iterative way to minimize the above function and the adversarial perturbation of TFA is updated as follows.
\begin{equation}
\begin{aligned}
\bm{\delta}_t = Clip\left(\bm{\delta}_{t-1} - \alpha \times sign\left( \cfrac{\partial\mathcal{L}}{\partial\bm{\delta}_{t-1}}   \right),-\epsilon,\epsilon\right),
\end{aligned}
\label{e8}
\end{equation} 
where $\alpha$ is  a fixed step size during optimization, $sign(\cdot)$ is the sign function, the function $Clip\left(\bm{\delta},-\epsilon,\epsilon\right)$ is used to restrict the value of $\bm{\delta}$ to the range of $ \left(-\epsilon,\epsilon\right)$.
After a number of  training iterations, the adversarial example related to the targeted object can be generated as, 
\begin{equation}
\begin{aligned}
\textbf{x}^{adv} = Clip(\textbf{x}^s + \bm{\delta}, 0 , 255).
\end{aligned}
\label{e9}
\end{equation} 

Similar to~\cite{2020Understanding}, the proposed TFA algorithm can be used to generate both image-specific perturbation and universal (image-agnostic) perturbation.
The image-specific perturbation~\cite{gao-et-al:ijcai2} is generated on a certain image and can only attack this image, while the universal perturbation~\cite{gd-uap,image_specific} is generated on an image set and can attack most images of this set.
The detailed generation processes are shown in Algorithm~\ref{alg:alg1} and Algorithm~\ref{alg:alg2}, respectively.

\begin{algorithm}[!h]
\caption{ Generating image-specific perturbation for object detection.}\label{alg:alg1}
\begin{algorithmic}
\STATE $ \textbf{Input: }  $ victim image  $\textbf{x}^s$, guided image  $\textbf{x}^g$, object detector $G$, restriction $\epsilon$, step size $\alpha $
\STATE  $ \textbf{Output: }  $ adversarial example $\textbf{x}^{adv}$
\STATE  $\textbf{v}^g \gets G(\textbf{x}^g)$ 
\STATE  Compute class attention $\textbf{w}^c$ using Equation (\ref{e4})
\STATE  Compute spatial attention $\textbf{w}^s$ using Equation (\ref{e5})
\STATE  Compute dual attention $\textbf{w}$ using Equation (\ref{e6})
\STATE $\textbf{x}^{adv} \gets \textbf{x}^s $
\STATE  \textbf{for} t $\gets$ 0 \textbf{to} $max_{iter}$ \textbf{do}
\STATE \hspace{0.5cm} $\textbf{v}_t^{adv} \gets G(\textbf{x}^{adv})$ 
\STATE \hspace{0.5cm} compute loss $\mathcal{L}$ using Equation (\ref{e7})
\STATE \hspace{0.5cm} $\bm{\delta}_t \gets Clip\left(\bm{\delta}_{t-1} - \alpha \times sign\left( \cfrac{\partial\mathcal{L}}{\partial\bm{\delta}_{t-1}}   \right),-\epsilon,\epsilon\right)$

\STATE \hspace{0.5cm} $\textbf{x}_{t}^{adv} \gets Clip(\textbf{x}_{t-1}^{adv} + \bm{\delta}_t, 0, 255).$
\STATE \textbf{return } $\textbf{x}_t^{adv}$

\end{algorithmic}
\label{alg1}
\end{algorithm}

\begin{algorithm}[!h]
\caption{ Generating universal perturbation for object detection.}\label{alg:alg2}
\begin{algorithmic}
\STATE $ \textbf{Input: }  $ victim image set $\mathcal{X}^s$, guided image  $\textbf{x}^g$, object detector $G$, restriction $\epsilon$, step size $\alpha $
\STATE  $ \textbf{Output: }  $ adversarial examples set $\mathcal{X}^{adv}$, universal perturbation $\bm{\delta} $
\STATE  $\textbf{v}^g \gets G(\textbf{x}^g)$ 
\STATE  Compute class attention $\textbf{w}^c$ using Equation (\ref{e4})
\STATE  Compute spatial attention $\textbf{w}^s$ using Equation (\ref{e5})
\STATE  Compute dual attention $\textbf{w}$ using Equation (\ref{e6})
\STATE $\textbf{x}^{adv} \gets \textbf{x}^s $
\STATE  Initialize universal perturbation $\bm{\delta}_0 $
\STATE  \textbf{for} t $\gets$ 0 \textbf{to} $max_{iter}$ \textbf{do}
\STATE \hspace{0.5cm}  \textbf{for} $\textbf{x}^s$  in  $\mathcal{X}^s$  \textbf{do}
\STATE \hspace{0.5cm} \hspace{0.5cm} $\textbf{x}_t^{adv} \gets Clip(\textbf{x}^{s} + \bm{\delta}_{t-1},0,255) $
\STATE \hspace{0.5cm} \hspace{0.5cm}  $\textbf{v}^{adv} \gets G(\textbf{x}_t^{adv})$ 
\STATE \hspace{0.5cm} \hspace{0.5cm} compute loss $\mathcal{L}$ using Equation (\ref{e7})
\STATE \hspace{0.5cm} \hspace{0.5cm} $\bm{\delta}_t \gets Clip\left(\bm{\delta}_{t-1} - \alpha \times sign\left( \cfrac{\partial\mathcal{L}}{\partial\bm{\delta}_{t-1}}   \right),-\epsilon,\epsilon\right)$
\STATE  \textbf{for} $\textbf{x}^s$  in  $\mathcal{X}^s$  \textbf{do}
\STATE  \hspace{0.5cm} $ \mathcal{X}^{adv}$ append $      Clip(\textbf{x}^{s} + \bm{\delta}_{t},0,255)  $
\STATE \textbf{return } $\mathcal{X}^{adv}$, $\bm{\delta}_{t}$
\end{algorithmic}
\label{alg2}
\end{algorithm}

\newpage
\section{Experimental Results}
In this section, we carry out extensive experiments to demonstrate the effectiveness of the proposed TFA method. Specifically, the following contents mainly contain the experimental comparisons of two aspects.
(1) Subsection~\ref{method_comparison}—the attack performance of baselines and the proposed TFA are evaluated in the cases of image-specific, universality, and generalization attack scenarios. (2) Subsection~\ref{impact_factors}—We analyze the impacts of the perturbation degree, dual attention module, and guided image on the performance of TFA individually.

\subsection{Experimental Settings}
\subsubsection{Dataset and Metric}
The experiments in this paper are conducted on MS COCO dataset~\cite{lin-et-al:coco} and BDD100K dataset~\cite{bdd100k}.
MS COCO dataset is a large-scale dataset for multiple computer vision tasks. 
For object detection, MS COCO has the annotations of bounding boxes with 80 object categories.
We randomly select 1000 samples as the victim images and 8 target objects in different classes to conduct the targeted attack experiments.
For each class, we randomly choose one sample that contains the corresponding target object as the guided image.
In addition to the MS COCO dataset, we also evaluate the proposed attack method on the BDD100K dataset.
BDD100K is a large-scale diverse driving dataset that has the annotations of bounding boxes with 10 object categories. 
Similarly, we randomly select 1000 samples as the victim images and 6 target objects in different classes. 
For each class, we also choose one sample that contains the corresponding target object as the guided image.
Fig. 3 shows part of the guided images containing the target objects.

\begin{figure}[!t]
\setlength{\abovecaptionskip}{-0.2cm}
	\centering
	\includegraphics[width=0.8\linewidth]{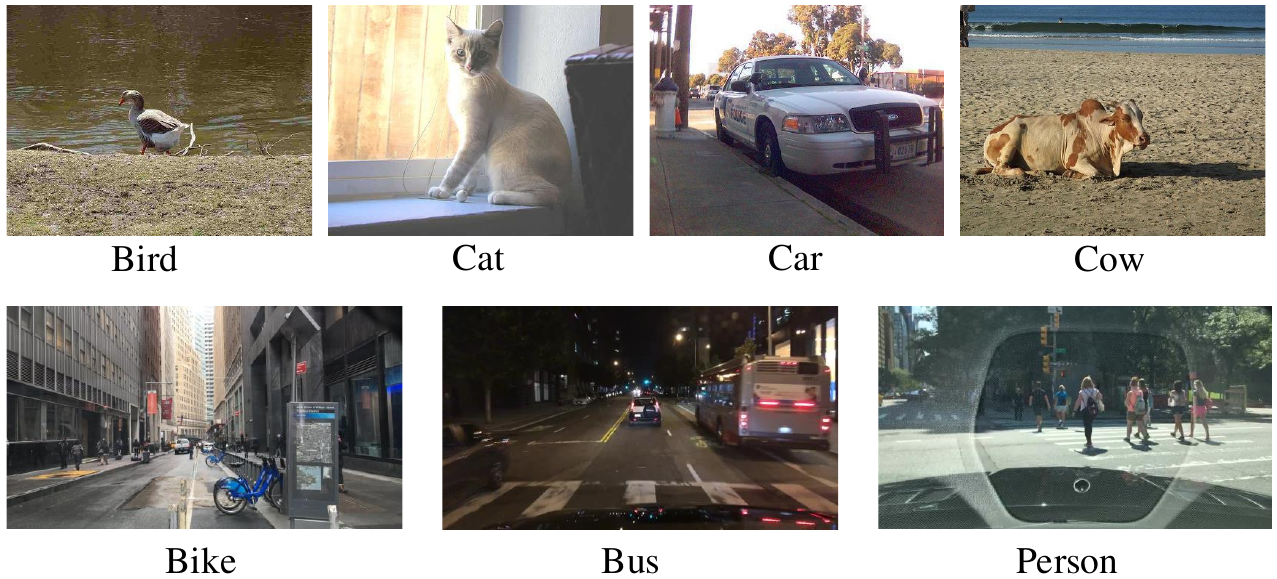}
	\caption{Part of the guided images containing the target objects. The first row is the guided images on the MS COCO dataset, second row is the guided images on the BDD100K dataset.}
	\label{fig:guide_img}
\end{figure}

Following~\cite{cai}, we use the success rate (SR) to evaluate the performance of attack methods.    
It is defined as the percentage of successful attacks that the adversarial example obtained after training can precisely mislead the detector to predict the target object.
Concretely, SR is calculated as,
 \begin{equation}
 \begin{aligned} 
SR = \cfrac{ Number~of~success~attacks}{Number~of~all~attacks}.
\end{aligned}
\label{e10}
\end{equation} 
During the experiments, each sample of the 1000 victim images will be attacked with the 8 selected target objects for the MS COCO dataset and 6 target objects for the BDD100K dataset individually, thus we present SR for each class in the following results. 

\subsubsection{Object Detectors}

To evaluate the attack performance on different object detectors, we exploit FasterRCNN and YOLOv5 to conduct the attack experiments, since they are the most representative detectors for two-stage detectors and one-stage detectors respectively.
To ensure the detectors are well-trained, we use the default parameters released in the official version \footnote{https://github.com/pytorch/vision} \footnote{https://github.com/ultralytics/yolov5}.
The confidence threshold for all detectors to output the detection results is set to 0.5.

\subsubsection{Comparison Methods} 
As introduced in Section~\ref{sec:relate}, all the existing targeted attack methods attack the final output layers to mislabel the detected objects and cannot be directly applied to the images without objects. To make a comprehensive comparison, we modify the representative method TOG and CAA to implement the attack. Specifically, the original TOG and CAA maximize the classification score of the specific wrong label to mislabel one detected object, while we modify it to optimize the objectness score and classification score of the final output layer simultaneously to ‘fabricate’ the designated target object.

\subsubsection{Hyperparameter Settings}
The proposed framework involves a few hyperparameters, they are configured with the same values for all methods. Specifically, the restriction value of the adversarial perturbation $\epsilon$ is set to 16, the number of iterations for one adversarial example is set to 80, and the step size $\alpha$ is set to 1.

\subsubsection{Attack Scenarios}

The proposed TFA method aims to improve the white-box attack performance in terms of image-specific, universality, and generalization attack scenarios for the targeted attack of object detection.
The explicit definitions for these aspects are shown below.

\begin{itemize}
\item 
\emph{Image-specific attack}: Given an image and a detector, the attacker conducts the optimization process to find an image-specific perturbation that can only be applied to the given image to attack the given detector.

\item
\emph{Universality attack}: Given a set of images and a detector, the attacker conducts the optimization process to find a universal perturbation that can be applied to most images of this set to attack the given detector.

\item
\emph{Generalization attack}: Given a set of new images that are not used to generate perturbation, the attacker directly applies the generated universal perturbations to these unseen data to attack the given detector.
\end{itemize}

We note that the metric for the above attack scenarios is the attack success rate defined in Eq.(\ref{e10}).

\subsection{Performance Comparisons }
\label{method_comparison}

\subsubsection{Image-specific Attack Performance}
The image-specific adversarial examples are generated by Algorithm~\ref{alg:alg1} and the attack success rate of each specific class is shown in Table~\ref{tab:white_data_COCO} and Table~\ref{tab:white_data_BDD}. 
We can see that the proposed TFA achieves significant improvements over TOG  and CAA in terms of the success rate for all the target objects.
Concretely, on the MS COCO dataset, the average success rates of TFA are 42.1 and 99.8 for attacking FasterRCNN and YOLOv5, while the average success rate of TOG/CAA is only 17.2/21.6 and 0.2/26.9, respectively.
On the BDD100K dataset, TFA achieves 90.5 and 98.3 average success rates for attacking FasterRCNN and YOLOv5 while TOG and CAA only achieve 60.9/69.4 and 0.2/24.4, respectively.
These results clearly demonstrate the effectiveness of our proposed TFA attack method.

\begin{table*}[t]
    \scriptsize
    \setlength{\belowcaptionskip}{0.1cm}
	\renewcommand{\arraystretch}{1.2}
	\centering
		\caption{The image-specific attack performance comparisons on the MS COCO dataset.}
	\begin{tabular}{ c c c c c c c c c c c c }
		\hline
		Detector & Method 
		& bird & cat & car & cow & dog & laptop & motor & teddy & average \\
		\hline
		
		\multirow{3}{*}{FRCNN} 
		& TOG~\cite{chow-et-al:tog}  &17.9  & 29.3 & 22.6 & 6.8 & 26.6 & 13.6 & 6.5 & 13.9 & 17.2 \\  
        & CAA~\cite{cai}  &20.1  & 32.4 & 23.9 & 10.8 & 33.1 & 21.7 & 10.9 & 20.1 & 21.6 \\  
		&  TFA & \textbf{40.4}  & \textbf{55.7} & \textbf{35.9} & \textbf{27.2} & \textbf{46.3} & \textbf{50.4} & \textbf{48.7} & \textbf{32.3}  & \textbf{42.1} \\
		\hline
		
		\multirow{3}{*}{YOLOv5} 
		& TOG~\cite{chow-et-al:tog}   & 0.4  & 0.1 & 0.0 & 0.0 & 0.8 & 0.0 & 0.0 & 0.1 & 0.2 \\  
        & CAA~\cite{cai}   & 21.4  & 19.4 & 25.1 & 33.7 & 26.3 & 30.2 & 28.1 & 31.2 & 26.9 \\  
		& TFA  &\textbf{99.8}  & \textbf{98.4} &\textbf{99.9} & \textbf{100.0} & \textbf{100.0} & \textbf{100.0} & \textbf{100.0} & \textbf{99.9} & \textbf{99.8} \\  		
		\hline
	\end{tabular}

	\label{tab:white_data_COCO}
 \vspace{-10pt}
\end{table*}

\begin{table*}[t]
    \scriptsize
	\renewcommand{\arraystretch}{1.2}
	\setlength{\belowcaptionskip}{0.1cm}
	\centering
	\caption{The image-specific attack performance comparisons on the BDD100K dataset.}
	\begin{tabular}{c c c c c c c c c }
		\hline
		Detector & Method 
		& bike & bus& motor& person& rider & truck & average \\
		\hline
		\multirow{3}{*}{FRCNN} 
		& TOG~\cite{chow-et-al:tog}   & 55.7 & 74.0 & 56.0 & 55.3 & 46.9 & 77.2 & 60.9 \\ 
         & CAA~\cite{cai}   & 65.2 & 80.3 & 62.9 & 66.2 & 61.5 & 80.1 & 69.4 \\ 
		& TFA &\textbf{99.6} &\textbf{98.1} &\textbf{90.5}  &\textbf{86.1} &\textbf{72.1} &\textbf{96.5} &\textbf{90.5}\\
		\hline
		\multirow{3}{*}{YOLOv5} 
		& TOG~\cite{chow-et-al:tog}  & 0.0 & 0.2 & 0.0 & 0.1 & 0.0 & 0.6 & 0.2\\ 
        & CAA~\cite{cai}  & 20.1 & 19.6 & 25.3 & 30.8 & 27.2 & 23.4 & 24.4\\ 
		& TFA  &\textbf{100.0} &\textbf{98.8} &\textbf{99.6} &\textbf{100.0}&\textbf{91.5 }&\textbf{99.9}&\textbf{98.3}\\ 
		\hline
	\end{tabular}
	\label{tab:white_data_BDD}
 \vspace{-15pt}
\end{table*}

An interesting phenomenon shown in Table~\ref{tab:white_data_COCO} and~\ref{tab:white_data_BDD} is that TOG seems to be more specialised in attacking FasterRCNN while TFA is more specialised in attacking YOLOv5.
On MS COCO and BDD100K datasets, the success rates of TOG on YOLOv5 are close to zero while the success rates of TOG for FasterRCNN are 17.2 and 60.9 respectively. As mentioned above, TOG attacks the final output layer by maximizing the classification score and objectness score. For the one-stage detector YOLOv5,  the classification and objectness are calculated in the same output layer. We argue that the optimization directions of these two components  may be inconsistent, thus leading to a non-convergence results and low success rate. In contrast, the  two-stage detector FasterRCNN outputs the  classification and objectness in different output layers, and they can be attacked and optimized separately to generate adversarial examples. For the proposed TFA, the success rates of YOLOv5 are higher than that of FasterRCNN on both datasets. We conjecture that this is mainly because the features extracted by YOLOv5 are more vulnerable to adversarial attack than that of FasterRCNN. The following experiments with respect to the perturbation degree in Fig.~\ref{fig:noise_degree} also show that the features extracted by YOLOv5 can be easily changed by TFA even with small $\epsilon$.

\begin{figure}[]
\setlength{\abovecaptionskip}{-0.2cm}
	\centering
	\includegraphics[width=0.9\linewidth]{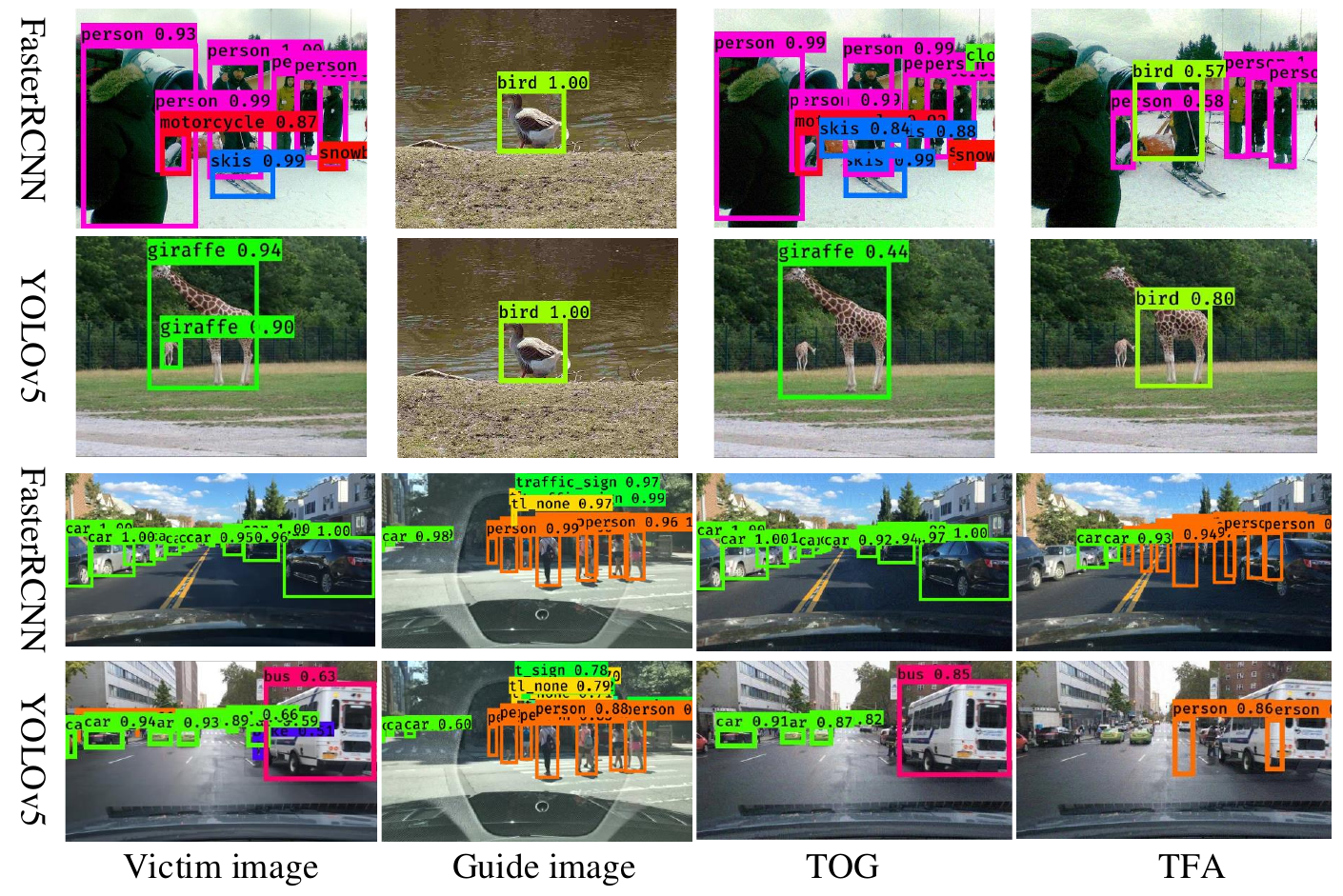}
	\caption{ The subjective illustrations of the proposed TFA and TOG on FasterRCNN and YOLOv5. The first two rows are the detection results of the MS COCO dataset while the last two rows are the detection results of the BDD100K dataset. Better to zoom in electronic version for viewing. }
	\label{fig:white_vis}
\end{figure}

\begin{figure}[]
	\centering
	\includegraphics[width=0.8\linewidth]{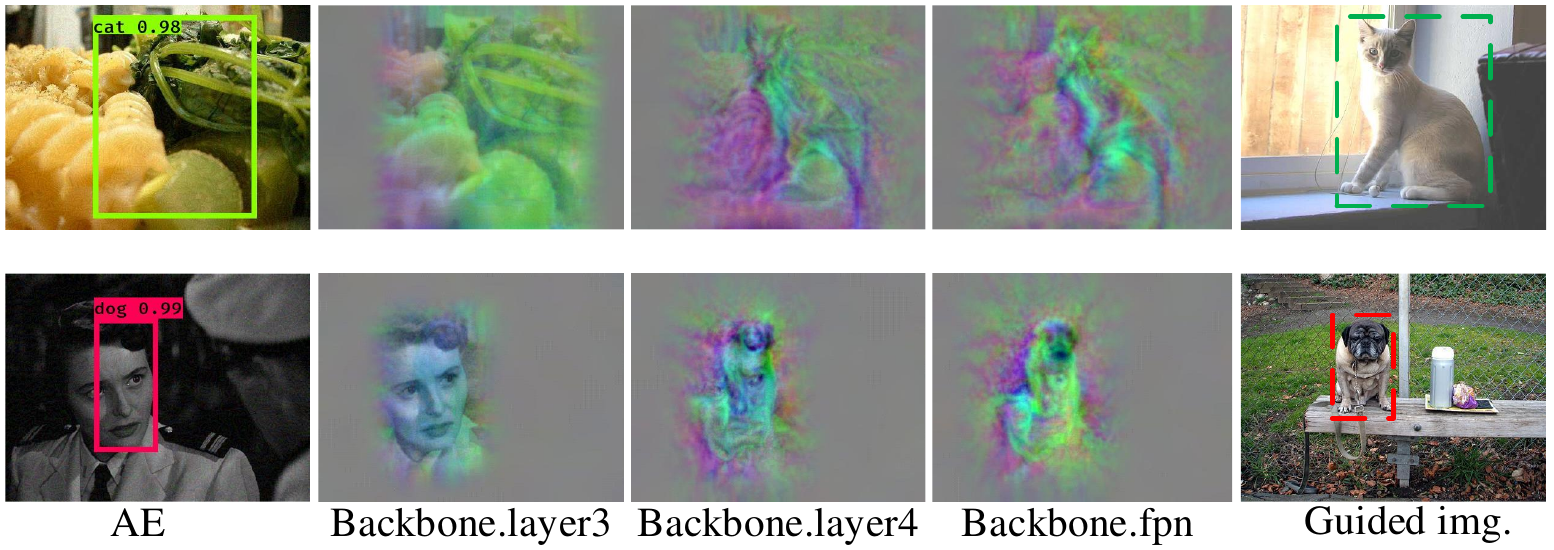}
	\caption{ The semantic information changes of different layers of FasterRCNN when performing TFA on two victim images.}
	\label{fig:segmantic}
\end{figure}

Fig.~\ref{fig:white_vis} further illustrates the subjective detection results of the proposed TFA and TOG on FasterRCNN and YOLOv5. 
The first column is the detections of the victim images without any perturbations. 
The second column is the detections of the guided images containing the target objects (e.g. bird and person).
The third and fourth columns are the detections of the adversarial examples generated by TOG and our TFA. 
Obviously, TOG fails to detect the targeted objects in these samples, and TFA can mislead the detectors to incorrectly detect the specified class at a certain image region where there is not even any object.

Fig.~\ref{fig:segmantic} illustrates the semantic information changes of different layers of FasterRCNN when performing TFA on two victim images. 
By employing the feature inversion technique~\cite{Mahendran-et-al:invert}, we can see that as the layers progress from shallow to deep, the semantic features of the target objects (e.g., cat and dog) are gradually transferred to the feature maps of the victim images. 
This demonstrates that the feature space is vulnerable to perturbations and can be changed purposely.

\subsubsection{Universality Attack Performance}
The universal perturbations and the corresponding adversarial examples are generated by Algorithm~\ref{alg:alg2}, and the universality attack results of TOG and TFA on two datasets are shown in Table~\ref{tab:universal_coco} and Table~\ref{tab:universal_bdd}. As it can be observed, the attack success rates of TFA are much higher than that of TOG and CAA for each specific class. This demonstrates that an elaborate targeted feature space attack has a better universal performance than the method attacking the final output layer. 
Similar to the image-specific attack, the success rates of TOG on YOLOv5 are close to zero in the case of the universality attack.
We also find that the average success rates on BDD100K dataset are higher than the MS COCO dataset, since the images in BDD100K are mainly driving scenarios and the features are similar in different images.

\begin{table*}[!t]
\scriptsize
    \setlength{\belowcaptionskip}{0.1cm}
	\renewcommand{\arraystretch}{1.2}
	\centering
		\caption{The universality attack performance comparisons on MS COCO dataset.}
	\begin{tabular}{ c c c c c c c c c c c c }
		\hline
		Detector & Method 
		& bird & cat & car & cow & dog & laptop & motor & teddy & average \\
		\hline
		\multirow{3}{*}{FRCNN} 
        & TOG~\cite{chow-et-al:tog}    & 32.0 & 18.3 & 50.0 & \textbf{32.0} & 13.9 & 10.0 & 14.8 & 20.5 & 23.9\\
        &  CAA~\cite{cai}  &  35.6 & 22.3 & 52.8 & 29.4 & 21.0 & 19.2 & 23.1 & 28.3 & 29.0\\
		& TFA  &\textbf{44.3} &\textbf{55.1}& \textbf{57.2} & 25.8 & \textbf{40.1} & \textbf{49.9} &\textbf{25.0}&\textbf{35.3} & \textbf{41.6} \\
		\hline
		
		\multirow{3}{*}{YOLOv5}
		& TOG~\cite{chow-et-al:tog}   & 0.3  & 1.1 & 0.1 & 0.0 & 0.0 & 0.0 & 0.0 & 0.1 & 0.2  \\  
         & CAA~\cite{cai}   & 26.3  & 32.5 & 22.5 & 19.9 & 31.1 & 28.2 & 30.2 & 27.6 & 27.3  \\  
		 & TFA  & \textbf{88.1} & \textbf{77.8} & \textbf{91.7} & \textbf{88.1} & \textbf{84.4} & \textbf{88.0} & \textbf{81.9} & \textbf{80.3} & \textbf{85.0}  \\  		
	\hline
	\end{tabular}
 \vspace{-10pt}

	\label{tab:universal_coco}
\end{table*}

\begin{table*}[]
\scriptsize
    \setlength{\belowcaptionskip}{0.1cm}
	\renewcommand{\arraystretch}{1.2}
	\centering
	\caption{The universality attack performance comparisons on BDD100K dataset.}
	\begin{tabular}{c c c c c c c c c }
		\hline
		Detector & Method 
		& bike & bus& motor& person& rider & truck & average \\
		\hline
		\multirow{3}{*}{FasterRCNN~\cite{ren-et-al:fasterrcnn}} 
		& TOG~\cite{chow-et-al:tog}  & 65.4 & 80.7 & 60.4 & 72.3 & 53.3 & 73.3 & 67.6 \\ 
  & CAA~\cite{cai}  & 70.6 & 83.1 & 70.9 & 76.1 & 59.9 & 78.2 & 73.1 \\ 
		& TFA  &\textbf{97.3 } &\textbf{85.6 }&\textbf{ 83.0} &\textbf{99.8} &\textbf{75.5} &\textbf{ 90.0}&\textbf{ 88.5}\\ 
		
		\hline
		\multirow{3}{*}{YOLOv5~\cite{yolov5}}
		& TOG~\cite{chow-et-al:tog}   & 0.0 & 1.0 & 0.3 & 0.1 & 0.0 & 2.6 & 0.7\\ 
          
        & CAA~\cite{cai}   & 21.4 & 19.3 & 22.0 & 33.8 & 26.7 & 29.6 & 25.5\\     
		& TFA &\textbf{99.4}  &\textbf{95.5 }&\textbf{97.6} &\textbf{100.0} &\textbf{88.6} &\textbf{ 96.4}&\textbf{96.3}\\  
	    \hline
	\end{tabular}
	\label{tab:universal_bdd}
\end{table*}

\begin{figure}[]
	\setlength{\abovecaptionskip}{-0.2cm}
	\centering
	\includegraphics[width=0.8\linewidth]{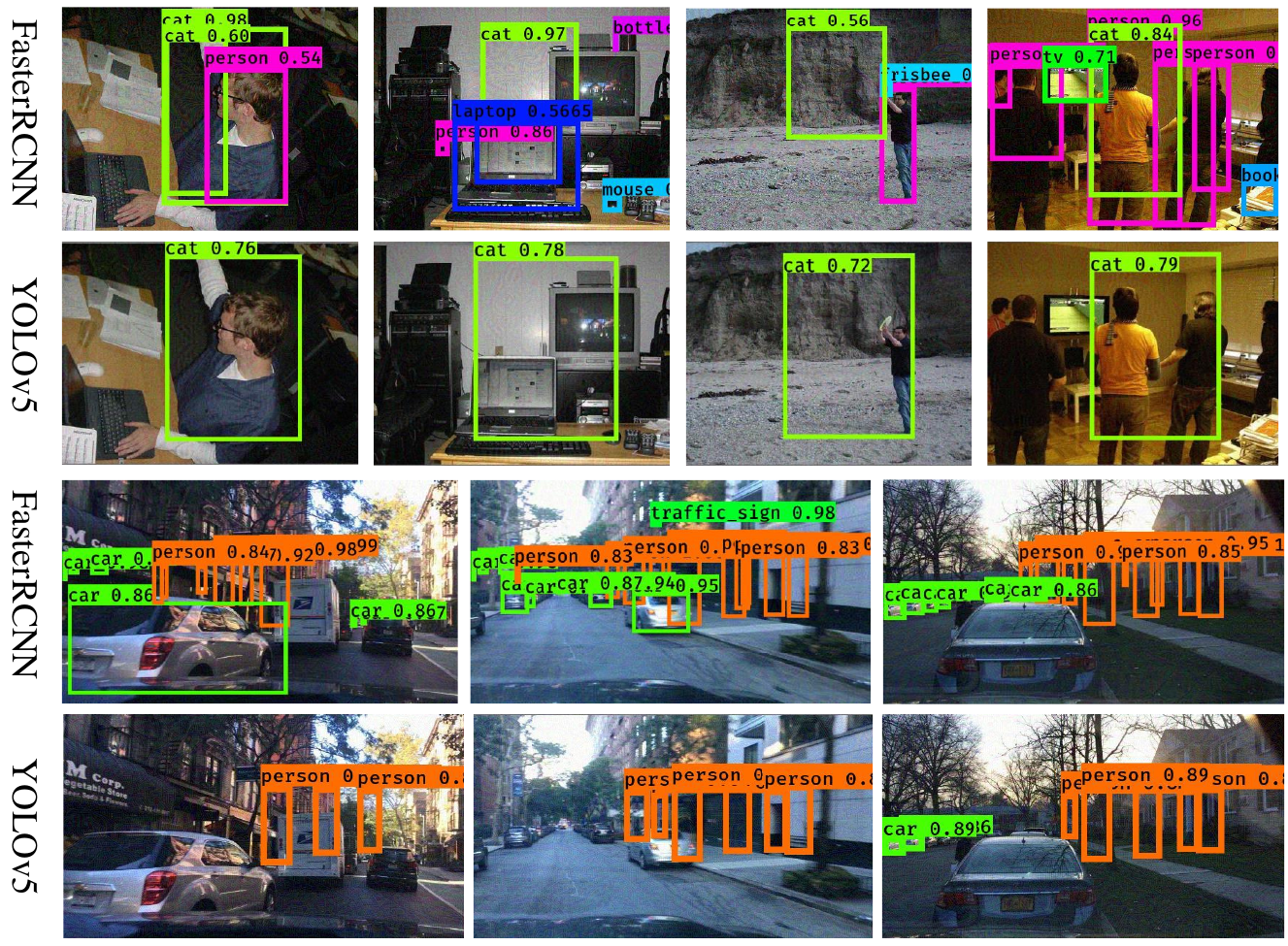}
	\caption{The subjective illustrations of the universality attacks (cat and person) on FasterRCNN and YOLOv5. The first two rows are the detection results of the MS COCO dataset while the last two rows are the detection results of the BDD100K dataset. Better to zoom in electronic version for viewing. }
 \vspace{-10pt}
	\label{fig:universal}
\end{figure}

\begin{figure}[]
\setlength{\abovecaptionskip}{-0.2cm}
	\centering
	\includegraphics[width=0.8\linewidth]{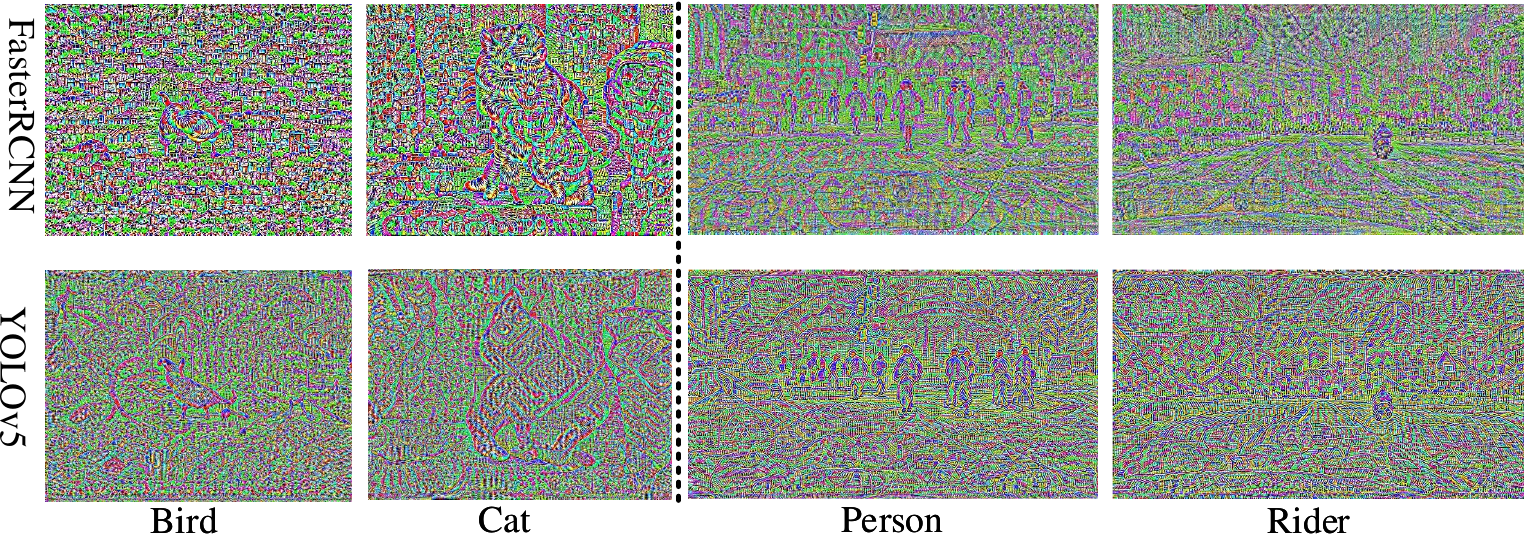}
	\caption{The visualization results of the universal perturbations for the target objects bird, cat, person, and rider. The first two columns are generated on the MS COCO dataset while the last two columns are generated on the BDD100K dataset.}
	\label{fig:universal_show}
\end{figure}

Fig.~\ref{fig:universal_show} shows the visualization results of the universal perturbations for the target objects (e.g., bird, cat, person, and rider).
In order to visualize these perturbations more clearly, we normalize the values from  (-16, 16) to (0, 255).
The first and second rows are the generated universal perturbations using FasterRCNN and YOLOv5 respectively. We can see that the general outlines of these target objects can be recognized in the universal perturbations, which implies the proposed method precisely locates the critical  features of the target objects from the guided images. Then, the perturbations are added to different victim images to achieve the universality attack.  Fig.~\ref{fig:universal} illustrates some subjective detection results of MS COCO and BDD100K datasets by applying the universal perturbations of cat and person on FasterRCNN and YOLOv5.

\subsubsection{Generalization Attack Performance}
\label{general}

In addition to the above image-specific attack and universality attack, we further employ the generated universal perturbations to attack unseen data to evaluate the generalization performance of  the proposed method. Specifically, we randomly select other 200 images from each of MS COCO and BDD100K datasets to construct the adversarial examples. It should be noted that these 200 images are not included in the victim image set $\mathcal{X}^s$ during the training process. 

The attack success rates with regard to the generalization performance are shown in Table~\ref{tab:generalization_performance_coco} and Table~\ref{tab:generalization_performance_bdd}.
On MS COCO dataset, TFA achieves 42.6 and 78.8 average success rates when attacking FasterRCNN and YOLOv5, while TOG/CAA only achieves 22.5/29.8 and 0.9/26.2 average success rates. 
Similar results occur on the BDD100K dataset, TFA achieves 89.3 and 97.4 average success rates while TOG/CAA only achieves 68.9/75.6 and 1.1/28.2 average success rates, respectively. These comparisons show that the proposed TFA has a stronger generalization performance than the previous methods.

Besides, we also observe that the success rates of the generalization performance of TFA in Table~\ref{tab:generalization_performance_coco} and Table~\ref{tab:generalization_performance_bdd} are comparable to the results in Table~\ref{tab:white_data_COCO} and Table~\ref{tab:white_data_BDD}. This means that the universal perturbations generated by the proposed method do contain the critical semantic information of the target objects and can effectively attack both the training data and unseen data.

\begin{table*}[!t]
\scriptsize
    \setlength{\belowcaptionskip}{0.1cm}
	\renewcommand{\arraystretch}{1.2}
	\centering
		\caption{
		The generalization attack performance comparisons between TOG and the proposed TFA on MS COCO dataset.
	}
	\begin{tabular}{ c c c c c c c c c c c c }
		\hline
		 Detector & Method 
		& bird & cat & car & cow & dog & laptop & motor & teddy & average \\
		\hline
		
		\multirow{3}{*}{FRCNN} 
	    	& TOG~\cite{chow-et-al:tog}  & 29.7 & 17.6 & 55.0 & \textbf{28.8} & 11.5 & 8.3 & 11.3 & 17.6 & 22.5
	     \\ 
      & CAA~\cite{cai}  & 31.2 & 28.6 & 60.3 & 27.9 & 23.4 & 17.8 & 23.9 & 25.0 & 29.8
	     \\ 
	     & TFA & \textbf{43.8} & \textbf{54.0} & \textbf{63.9} & 28.3 & \textbf{43.5} & \textbf{50.3} & \textbf{20.6} & \textbf{36.4} & \textbf{42.6}
	     \\
	     \hline
           \multirow{3}{*}{YOLOv5}
	    	&TOG~\cite{chow-et-al:tog}  & 1.0 & 3.7 & 1.1 & 0.0 & 0.0 & 0.0 & 0.0 & 1.0 & 0.9
	     \\ 
      &CAA~\cite{cai}   & 30.1 & 27.5 & 23.9 & 19.3 & 24.6 & 30.0 & 26.2 & 27.7 & 26.2
	     \\ 
      
	    & TFA & \textbf{85.3} & \textbf{69.6} & \textbf{92.6} & \textbf{76.8} & \textbf{78.9} & \textbf{85.2} & \textbf{71.4} & \textbf{70.5} & \textbf{78.8}
	     \\
	    \hline
	\end{tabular}
	\label{tab:generalization_performance_coco}
\end{table*}

\begin{table*}[!t]
\scriptsize
    \setlength{\belowcaptionskip}{0.1cm}
	\renewcommand{\arraystretch}{1.2}
	\centering
	\caption{The generalization attack performance comparisons between TOG and the proposed TFA on BDD100K dataset.}
	\begin{tabular}{c c c c c c c c c }
		\hline
		Detector & Method 
		& bike & bus& motor& person& rider & truck & average \\
		\hline
		\multirow{3}{*}{FasterRCNN~\cite{ren-et-al:fasterrcnn}}
		&TOG~\cite{chow-et-al:tog}  & 64.8 & 73.9 & 63.4 & 81.3 & 56.4 & 73.7 & 68.9
	     \\ 
      &CAA~\cite{cai}   & 73.0 & 82.9 & 70.5 & 85.2 & 62.0 & 79.8 & 75.6
	     \\ 
	    & TFA & \textbf{99.4} &\textbf{88.6}&\textbf{85.6}&\textbf{99.1}&\textbf{75.2}&\textbf{88.0}&\textbf{89.3}
	     \\
	    \hline
			\multirow{3}{*}{YOLOv5~\cite{yolov5}}
	     &TOG~\cite{chow-et-al:tog}  & 0.0 & 2.6 & 0.0 & 0.0 & 0.0 & 3.8 & 1.1
	     \\ 
      &CAA~\cite{cai}   & 31.2 & 29.6 & 21.6 & 29.4 & 32.0 & 25.1 & 28.2
	     \\ 
	     & TFA  & \textbf{99.5} & \textbf{96.9} & \textbf{97.9} & \textbf{100.0} & \textbf{92.0} & \textbf{98.1} & \textbf{97.4}
	     \\
	    \hline
	\end{tabular}
 \vspace{-10pt}
	\label{tab:generalization_performance_bdd}
\end{table*}

\subsection{Detailed Analysis on TFA}
\label{impact_factors}
 
\subsubsection{Ablation Study on Dual Attention}
\begin{table*}[!t] 
\scriptsize
    \setlength{\belowcaptionskip}{0.1cm}
\renewcommand{\arraystretch}{1.2}
\centering
\caption{The comprehensive ablation study results of the proposed dual attention mechanism on attacking FasterRCNN and YOLOv5.
}
\begin{tabular}{c c c c c c c c c c c c c}
\hline
 Detector & Attention 
      & bird & cat & car & cow & dog & laptop & motor & teddy  & average \\
\hline
\multirow{4}{*}{FRCNN}
& None   & 28.5  & 41.6 &30.0 & 14.4 & 37.1 & 36.9 & 15.2 & 6.7  & 26.3\\ 
& Class   & 37.6  & 44.4 & 35.4 & 15.4 & 40.6 & 42.9 & 25.0 & 14.3  & 32.0\\  
& Spatial   & 31.6  & 42.2 & 32.3 & 19.3 & 38.4 & 39.2 & 18.8 & 10.5  & 29.0\\  
& Dual   & \textbf{40.4}  & \textbf{55.7} & \textbf{35.9} & \textbf{27.2} & \textbf{46.3} & \textbf{50.4} & \textbf{48.7} & \textbf{32.3}  & \textbf{42.1}\\  
\hline
\multirow{4}{*}{YOLOv5}
& None       &94.9  &74.9   &92.7 &98.8 &89.0 &97.2 &92.9 &86.9 &90.9\\ 
& Class     &95.9  &83.6   &97.1 &99.7 &95.7 &97.4 & 94.5 & 94.1 &94.8\\ 
& Spatial   &96.1  &79.8   &95.3 &99.6 &96.1 &99.0 &95.8 &90.5 &94.0\\  
& Dual    &\textbf{96.2}   &\textbf{84.2}  &\textbf{98.3}  &\textbf{99.9} &\textbf{98.3} &\textbf{99.0} &\textbf{96.8} &\textbf{95.7} &\textbf{96.1}\\ 
\hline
\end{tabular}
\label{tab:white-ablation_fasterrcnn}
\vspace{-10pt}
\end{table*}

As presented in Section~\ref{sec:TFA},  the dual attention mechanism is a key component for the  proposed TFA because it can filter out the critical features of the target object in the whole feature space. 
Therefore, we conduct an ablation study to demonstrate the effectiveness of the class attention and the spatial attention. 
Specifically, the attack success rates of TFA on FasterRCNN and YOLOv5 are evaluated respectively in the cases of $\textbf{w}_i  =1$, $\textbf{w}_i  ={w}^c_i$, $\textbf{w}_i  =\textbf{w}^s_i$ and $\textbf{w}_i  ={w}^c_i \times \textbf{w}^s_i$. 
Table~\ref{tab:white-ablation_fasterrcnn} shows the results of the ablation study on MS COCO dataset,  where none attention corresponds to $\textbf{w}_i  =1$, class attention corresponds to $\textbf{w}_i  =\textbf{w}^c_i$,   spatial attention corresponds to $\textbf{w}_i  =\textbf{w}^s_i$,  and dual attention corresponds to $\textbf{w}_i  ={w}^c_i \times \textbf{w}^s_i$. We note that the results of YOLOv5 is obtained in the cast of $\epsilon$=8 for a clear comparison.

\begin{figure}[!t]
	\setlength{\abovecaptionskip}{-0.2cm}
	\centering
	\includegraphics[width=0.9\linewidth]{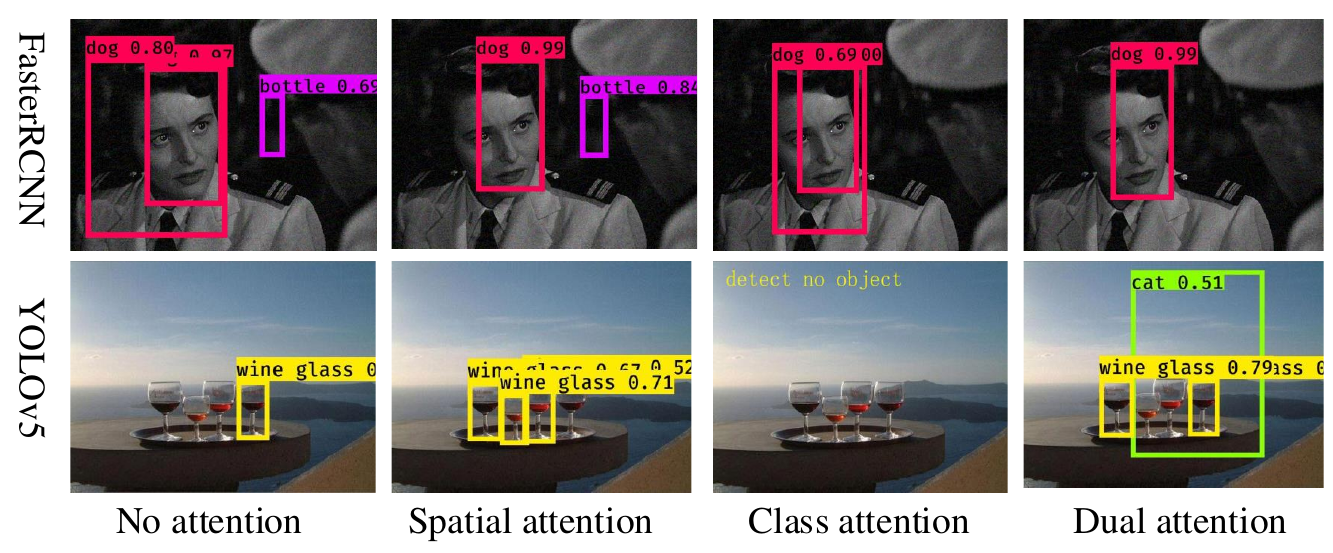}
	\caption{The subjective illustrations of the ablation study for TFA. The guided images and the target objects (i.e. dog and cat)  are shown in Fig.~\ref{fig:segmantic}.}
	\label{fig:ablation}
 \vspace{-10pt}
\end{figure}

As it can be observed, almost all the attack success rates of the none attention are the lowest because this case optimizes the whole feature space equally and does not focus on the critical features of the target objects.  
With the use of class attention or spatial attention, the success rate increases to some extent, and the dual attention gains the optimal performance. 
Fig.~\ref{fig:ablation} illustrates two samples of the corresponding subjective results for TFA. 
The first row is an example of attacking FasterRCNN with a dog as the target object, we can see that the detector predicts multiple objects other than the dog or predicts wrong positions in the first three columns, while the proposed dual attention ensures that the specific class  can  be detected uniquely in the right position. 
The second row is an example of attacking YOLOv5 with a cat as the target object, the detector can not predict the specific cat with only spatial attention or class attention, while the specific class can be fabricated successfully by employing the dual attention mechanism.

\subsubsection{Analysis of Perturbation Degree}
\label{perturbation_degree}
In the hyperparameter setting, the restriction value of the adversarial perturbation $\epsilon$ is fixed to 16. However, different  perturbation degrees have different attack effects for the targeted attack method. Table~\ref{tab:perturbation_results} shows the  comparison results of TFA when attacking FasterRCNN and YOLOv5 in the cases of $\epsilon$=4, 8, 12, 16 and 20 respectively. We can see that the success rates of TFA attack on FasterRCNN drop significantly as the decrease of $\epsilon$. When $\epsilon$ is set to 4, the corresponding success rate is close to zero. In contrast, 
the proposed method keeps high performance ($>$ 96) for the YOLOv5 attack as long as $\epsilon$ is greater than 8. The success rate is 48.3 even if $\epsilon$ is set to 4. This comparison demonstrates that FasterRCNN may be more sensitive to the adversarial perturbation and the extracted features of YOLOv5 can be easily changed even with small perturbation.

\begin{figure}[!t]
	\setlength{\abovecaptionskip}{-0.2cm}
	\centering
	\includegraphics[width=0.9\linewidth]{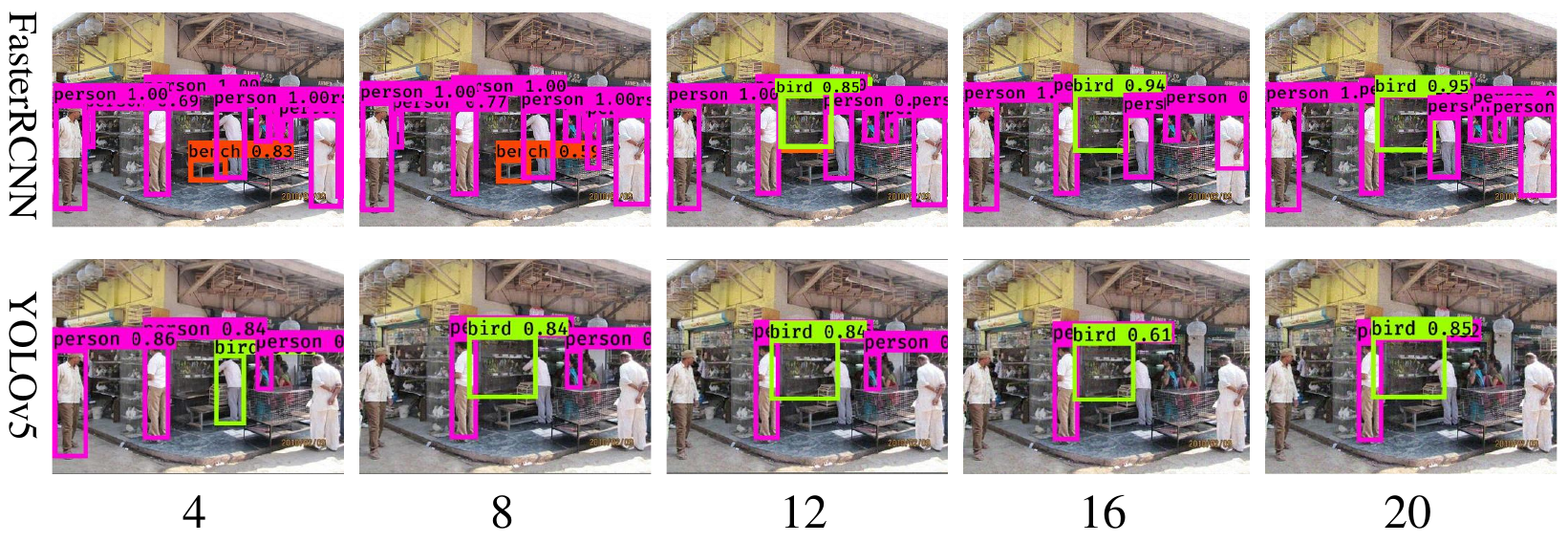}
	\caption{The visualization results of different perturbation degrees ($\epsilon$ = 4, 8, 12, 16, 20). The first and second rows are the detection results of TFA attack on FasterRCNN and YOLOv5 respectively. The target object is a bird.}
	\label{fig:noise_degree}
 
\end{figure}

Fig.~\ref{fig:noise_degree} illustrates the detection results of TFA at different $\epsilon$. The first and second rows are the visualization results of FasterRCNN and YOLOv5 for one image sample, respectively. The target object is a bird, and it can be observed that TFA successfully misleads YOLOv5 to predict a bird with various configurations of $\epsilon$ while the attacks on FasterRCNN fail in the cases of $\epsilon$=4 and $\epsilon$=8.

\begin{table}[!t]
\scriptsize
    \setlength{\belowcaptionskip}{0.1cm}
	\renewcommand{\arraystretch}{1.2}
	\centering
		\caption{The comparison results of TFA when attacking FasterRCNN and YOLOv5 in the cases of different $\epsilon$.}
	\begin{tabular}{c | c c c c c c c c c c c }
		\hline
		Detector & 4 & 8 & 12 & 16 & 20  \\
		\hline
		FRCNN &0.8  &8.0 &26.6 & 42.1 &51.0    \\  
		YOLOv5 & 48.3 & 96.1 & 99.6 & 99.8 & 99.9 \\  		
		\hline
		
	\end{tabular}

	\label{tab:perturbation_results}
\end{table}

\begin{figure}[t]
	\setlength{\abovecaptionskip}{-0.2cm}
	\centering
	\includegraphics[width=0.63\linewidth]{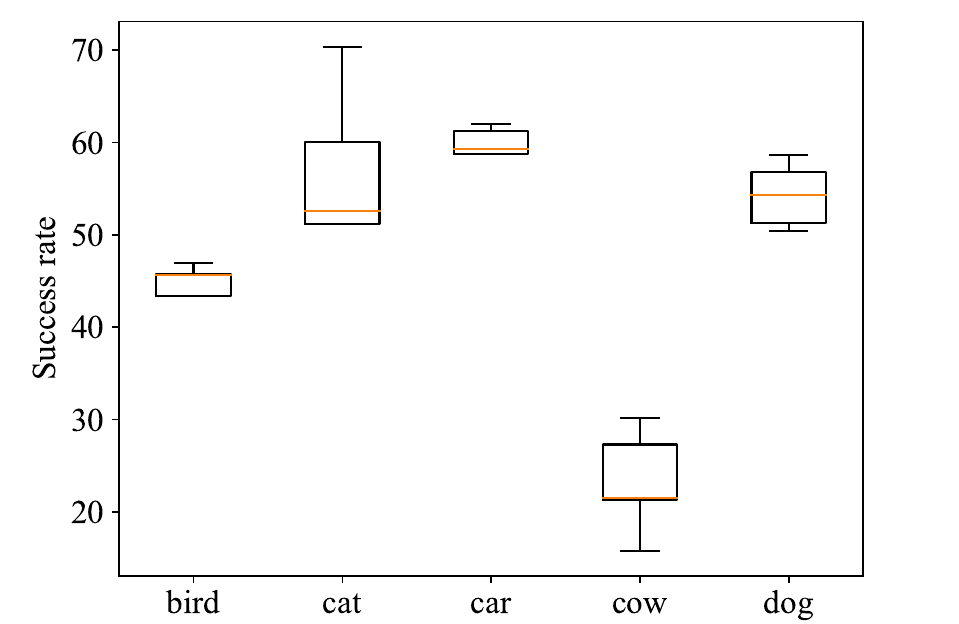}
	\caption{The success rates of different guided images for attacking FasterRCNN on MS COCO dataset.}
	\label{fig:guide_more}
 \vspace{-10pt}
\end{figure}

\begin{figure}[t]
	\setlength{\abovecaptionskip}{-0.2cm}
	\centering
	\includegraphics[width=0.8\linewidth]{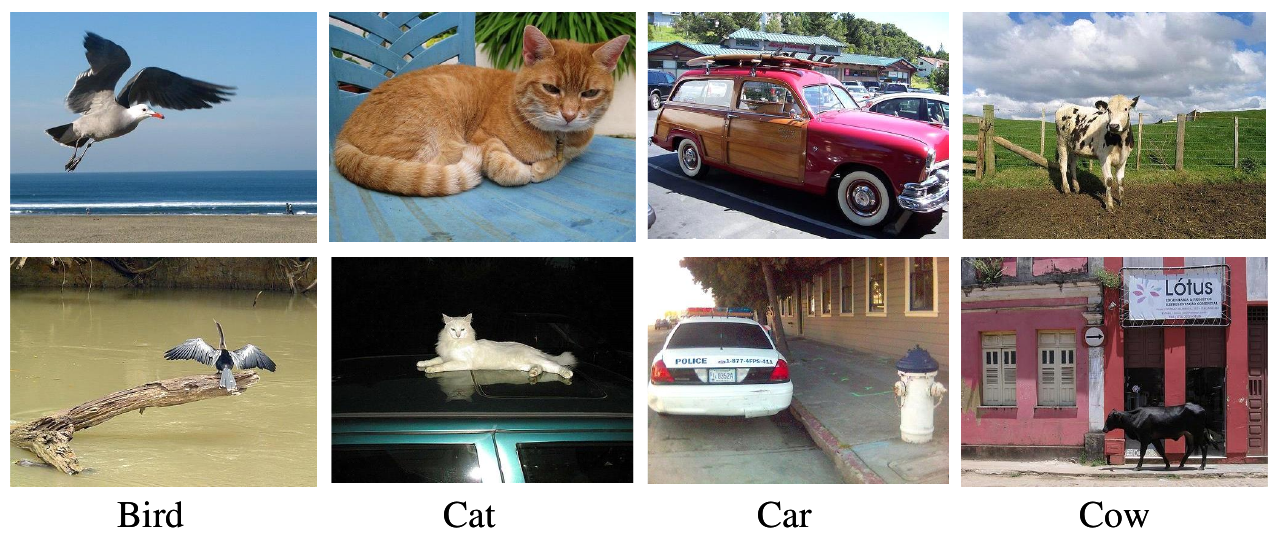}	\caption{The subjective illustrations of different guided images. The images in the first row achieve the highest success rates for the corresponding class while the images in the second row achieve the lowest success rates.}
	\label{fig:guide_high_low} 
\end{figure}
\subsubsection{Impact of Guided Image}

Guided image is another key component in the proposed TFA method as presented in Section~\ref{sec:TFA}. 
In the experimental setting, we randomly selected one sample containing the target object as the guided image. 
Therefore, it is necessary to evaluate the impacts of different guided images for one specific class. Fig.~\ref{fig:guide_more} illustrates the success rates of different guided images for attacking FasterRCNN on the MS COCO dataset. 
We can observe that different guided images of one specific class have distinctive attack performance. Taking the specific class dog as an example, the success rate varies from 50.4 to 58.7 for different guided images. Fig.~\ref{fig:guide_high_low} shows some guided images associated with the lowest and highest success rates of four target objects.
According to the experimental results and our analysis, we argue that the guided image including a complete and clear target object promotes the attack success rate.

The experiments on the impact of guided images show that the proposed TFA could further improve the attack performance if the guided images are assigned carefully rather than random selection. However, the above comparisons have shown a significant improvement of TFA over TOG even with the random selection.

\section{Conclusion}
This work proposes a new targeted attack method for object detection to mislead detectors to detect extra designated objects with specific target labels. we further design a novel attention based feature space attack method (TFA) that drives the extracted features of the victim images towards the target objects' features in the guided images, thus achieving the targeted attack of a specific class.
Extensive evaluations on multiple datasets (MS COCO and BDD100K) and detectors (FasterRCNN and YOLOv5) show that the proposed TFA method has better performance in terms of image-specific, universality, and generalization attack compared with previous works.
Currently, this paper focuses on the performance improvement of white-box attacks, we will study on the black-box setting of the targeted attack in future work.

\section{Acknowledgements}
This work was supported by the Fundamental Research Funds for the Central Universities [grant number xtr072022001];  National Natural Science Foundation of China [grant number 61790563].

\bibliographystyle{elsarticle-num} 
\bibliography{refs.bib}
\end{document}